\documentclass[11pt]{article}

\usepackage[preprint]{acl}

\usepackage{times}
\usepackage{latexsym}
\usepackage{amssymb}

\usepackage[T1]{fontenc}
\usepackage[utf8]{inputenc}

\usepackage{microtype}
\usepackage{float}

\usepackage{inconsolata}
\usepackage{linguex}
\AtBeginDocument{
  \setlength{\Exindent}{0pt}
  \setlength{\Exlabelwidth}{1.5em}
  \setlength{\Exlabelsep}{0.4em}
  \setlength{\SubExleftmargin}{1.2em}
}

\usepackage{CJKutf8}
\newcommand{\ja}[1]{\begin{CJK}{UTF8}{min}#1\end{CJK}}
\newcommand{\zh}[1]{\begin{CJK}{UTF8}{gbsn}#1\end{CJK}}

\newcommand{\dataset}{\textbf{CALQUE}}
\newcommand{\nsentences}{5583}

\usepackage{graphicx}
\usepackage{amsmath}
\usepackage{mathptmx}
\usepackage[table]{xcolor}
\definecolor{lightgreen}{HTML}{CDEDCD}
\definecolor{lightred}{HTML}{F5CACA}
\usepackage{booktabs}
\usepackage{multirow}
\usepackage{xfp}
\usepackage{url}

\newcommand{\failmin}{0}
\newcommand{\failmax}{80}
\newcommand{\failstrength}{35}
\newcommand{\failpct}[1]{\fpeval{round(min(\failstrength,max(0,\failstrength*(0#1-\failmin)/(\failmax-\failmin))),1)}}
\newcommand{\failcell}[1]{\cellcolor{red!\failpct{#1}!white}#1}

\title{\textit{I play of the cello}: \\[1mm]
Evaluating Constructional Bleeding in Multilingual LLMs with CALQUE}

\author{
\textbf{Toshiki Nakai}\textnormal{\textsuperscript{1}} \quad
\textbf{Yu-Yin Hsu}\textnormal{\textsuperscript{2}} \quad
\textbf{Marie-Catherine de Marneffe}\textnormal{\textsuperscript{3}}  \quad
\textbf{Joakim Nivre}\textnormal{\textsuperscript{4}} \quad \\
\textbf{Jakob Prange}\textsuperscript{5} \quad
\textbf{Nathan Schneider}\textsuperscript{6} \quad
\textbf{Wesley Scivetti}\textsuperscript{6} \quad 
\textbf{Tiago Timponi Torrent}\textsuperscript{7} \quad \\
\textbf{Leonie Weissweiler}\textsuperscript{1} \\
  \\
  \textsuperscript{1}Leipzig University \& ScaDS.AI Dresden/Leipzig \quad
  \textsuperscript{2}The Hong Kong Polytechnic University \quad \\
  \textsuperscript{3}CENTAL, UCLouvain \& FNRS \quad 
  \textsuperscript{4}Uppsala University \quad \\
  \textsuperscript{5}DZSKJ, University Medical Center Hamburg-Eppendorf \quad \\
  \textsuperscript{6}Georgetown University \quad
  \textsuperscript{7}Federal University of Juiz de Fora, FrameNet Brasil, CNPq \\
  \\
  \texttt{toshiki.nakai@uni-leipzig.de}
}

\begin{document}
\maketitle
\begin{abstract}
Languages use different morphosyntactic strategies to express the same meaning. Speakers of multiple languages are able to apply strategies from one language to another, either by mistake or intentionally, to produce a `literal' translation, or \textit{calque}. Considering that large language models (LLMs) have been shown to generalize crosslingually, despite being predominantly trained on English, we investigate whether they overgeneralize constructional strategies from English to other languages. To test this, we create a dataset of \nsentences{} sentence pairs across seven languages, matching literal translations of strategies from English into other languages, and literal translations of the other languages' strategies into English. We then quantify constructional bleeding by measuring how often strategies from English are assigned a higher probability than native strategies. We find that models often fail to assign higher probabilities to native strategies and tend to prefer English-derived strategies even when evaluating sentences in other languages. Our findings highlight the limitations of the current English-dominant training paradigm.\footnote{Data and Code:\\ \url{https://github.com/LionLabNLP/calque}} 
\end{abstract}

\section{Introduction}

Languages express the same meaning using systematically different forms. For example, in \ref{ex:cello}, English expresses the instrument as a direct object with a definite article (\textit{the cello}), while French introduces it with the preposition \textit{de}, fused with the definite article (\textit{du violoncelle}, from \textit{de} + \textit{le}).

\ex.\label{ex:cello}
\ag.\label{ex:cello-en}I play the cello.\\
\textsc{1sg} play \textsc{def} cello\\
\bg.\label{ex:cello-fr}Je joue du violoncelle.\\
\textsc{1sg} play.\textsc{1sg} of.\textsc{def} cello\\

% Such differences are not restricted to individual sentences. 
%\citet{croft2022morphosyntax} defines this difference as a \textit{strategy}:

%\begin{quote}
%    \textbf{strategy}: a construction in a language (or any
%language), used to express a particular combination of semantic content %and information packaging (the `what'), that is further distinguished
%by certain characteristics of grammatical form
%that can be defined in a crosslinguistically consistent fashion (the %`how') \cite[p.~22]{croft2022morphosyntax}
%\end{quote}

\noindent
The English and French constructions are language-specific, but they instantiate cross-lingually identifiable morphosyntactic patterns referred to as \emph{strategies} in linguistic typology 
\citep{keenan77,givon79,Croft+2016+377+393}. When using their native language, speakers can deploy these strategies productively. When using a second language, speakers sometimes produce literal translations that transfer a strategy from one language to another. This raises a fundamental question about multilingual language models: given how prevalent English is in the training data, \textbf{will models overgeneralize strategies found in English to other languages?}

\begin{figure}
    \centering
    \includegraphics[width=\linewidth]{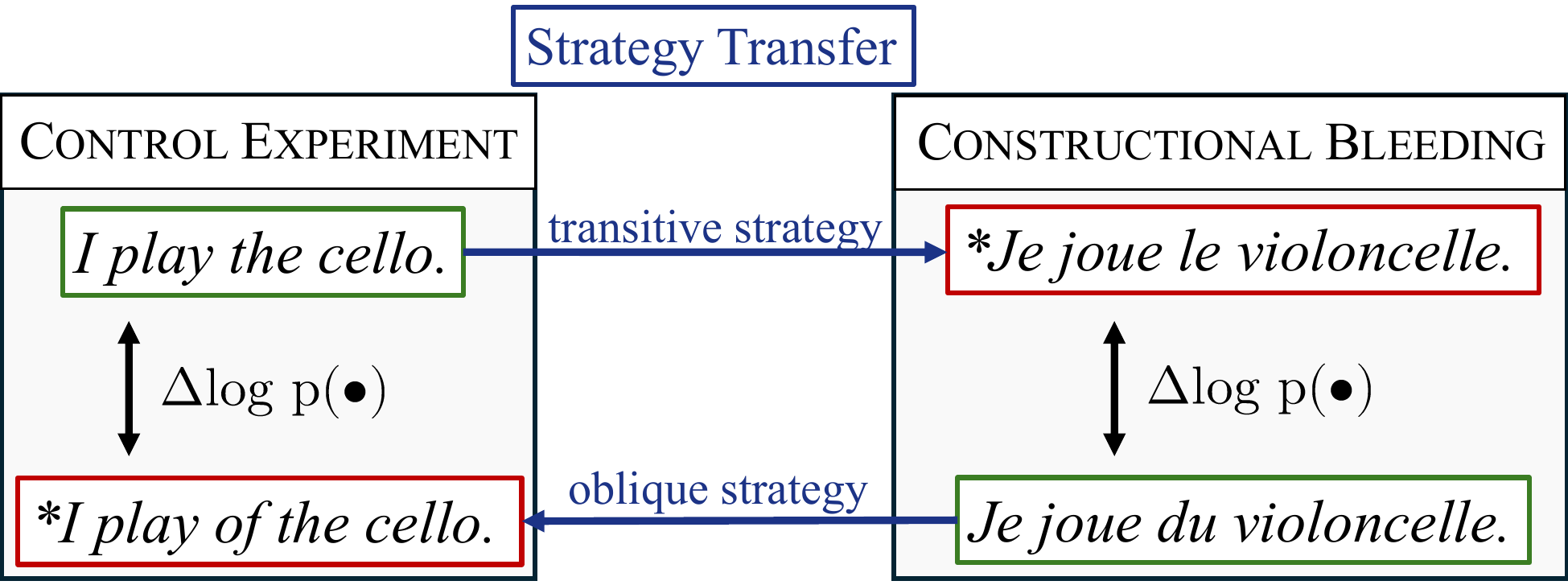}
    \caption{English and French speakers describe playing the cello with distinct morphosyntactic strategies. We measure \textit{bleeding} of English morphosyntax into French ($\Delta\log{p(\cdot)}$ between French sentences) and additionally conduct a control experiment on English ($\Delta\log{p(\cdot)}$ between English sentences).}
    \label{fig:page1}
\end{figure}

Interference from constructions in one language on the processing of another language can generally be described as \textbf{constructional bleeding}. Here we are interested in the special case when the interference occurs at the level of strategies, which we call \textbf{cross-linguistic strategy transfer}.   
We refer to a strategy conventionally used to express a given meaning in a language as the \textbf{native strategy}, and to a strategy transferred from another language as the \textbf{transferred strategy}. We introduce a controlled evaluation paradigm that holds semantic content approximately constant while varying the strategy used to express it. For each cross-linguistic contrast, as shown in Fig.~\ref{fig:page1}, we construct four sentences: an English sentence with the native strategy, an English sentence with the transferred strategy, a sentence in the other language with its native strategy, and a corresponding sentence with the English strategy transferred into that language. 

This design provides a controlled setting for studying cross-lingual strategy transfer. Multilingual LLMs are commonly trained on highly imbalanced multilingual corpora in which English constitutes a substantial share of the data, and previous work has found evidence of English-centric biases in their internal representations \citep{wendler,schut2025do,trinley-etal} and output behavior \citep{papadimitriou-etal-2023-multilingual,guo}. Do such biases afflict constructional strategy choices as well? By testing both directions of each strategy contrast, our framework allows us to examine whether strategy-level generalization is asymmetric: in particular, whether models overgeneralize strategies from English to other languages.

Within this framework, we introduce \dataset{}: \textbf{C}onstructional \textbf{A}lternations across \textbf{L}anguages: \textbf{Qu}antifying \textbf{E}nglish bias, a benchmark for evaluating constructional bleeding \textit{via} cross-lingual strategy transfer in multilingual LLMs. It covers seven languages, including English, and three broad construction types, with controlled contrasts between native and transferred strategies. Following the minimal-pair paradigm used in previous behavioral evaluations of language models \citep{warstadt-etal-2020-blimp-benchmark,hu-etal-2020-systematic,nangia-etal-2020-crows,misra-etal-2023-comps}, we evaluate model preferences using sentence probabilities. We benchmark 39 language models, including 28 English-centric models and 11 models with a stronger focus on non-English languages.

Using \dataset{}, we first confirm that models often struggle to distinguish native from transferred strategies, exhibiting lower failure rates in English than in other languages. Furthermore, the distributions of model preferences vary depending on their focus languages.

To control for differences in the evaluation language, we then evaluate models exclusively in Japanese, comparing native Japanese strategies with strategies transferred from the other six languages. We find evidence of asymmetric constructional bleeding: models exhibit greater difficulty distinguishing native Japanese strategies from English-transferred strategies than from strategies transferred from other languages.

Our main contributions are threefold:
(1)~\dataset{}, a controlled benchmark for evaluating language-specific constructional representation across \textbf{seven languages} and \textbf{three construction types};
(2)~results offering evidence for \textbf{constructional bleeding} in multilingual LLMs; and
(3)~a large-scale evaluation of 39 models establishing systematic differences between English-centric and non-English-centric models as well as strong correlations in strategy preferences across English-centric model families.

\section{Related Work}

\subsection{Crosslingual Transfer}

An analogy can be made to human acquisition of a second language. An L2 speaker will often model some aspect of their L2 output after their L1. In the field of second language acquisition, this is called \emph{transfer} \citep{jarvis2008crosslinguistic}. When transfer is facilitative (e.g., because the two languages share a strategy in common), it is regarded as a \emph{positive transfer}. When, however, it results in a formulation in the L2 that is ungrammatical or unidiomatic, it is a \emph{negative transfer} or \emph{interference} \citep{DellaPutta+2016+237+268}. Here we investigate a kind of negative transfer, albeit for multilingual models \citep[cf.~\unskip][]{yadavalli-etal-2023-slabert,oba-etal-2023-second,aoyama-schneider-2024-modeling} rather than humans.

A \emph{calque} is an expression resulting from hyper-literal translation \cite{Molina2004TranslationTR}.  We adopt it as the name of our dataset as the term evokes the sort of fluency contrasts in our evaluation (though we do not target a translation task).

\emph{Translationese} is another phenomenon whereby translated language differs in systematic ways from non-translated utterances or texts in the target language. It is often a statistical phenomenon rather than a qualitative one: for example, a translated text may contain a higher than expected number of connective words, even if each individual use of those connectives sounds perfectly natural. \citet{koppel-ordan-2011-translationese,jalota-etal-2023-translating,wein-schneider-2024} approach such a phenomenon from a computational perspective. Our study focuses on strategy-pairings where annotators find a strong qualitative difference in acceptability. Other methods would therefore be necessary to detect nuances associated with translationese. 

\subsection{Multilingual Representation and English Bias in LMs}\label{sec:eng-bias}

Multilingual language models display remarkable ability to transfer knowledge encoded in different languages \citep{pires-etal-2019-multilingual,conneau-etal-2020-unsupervised}. These models share internal representations across languages \citep{muller-etal-2021-first,chang-etal-2022-geometry,shani-basirat-2025-language}, though there is evidence that even shared internal representations are not fully language agnostic: typologically similar languages are more likely to share representations \citep{choenni-shutova-2022-investigating,stanczak-etal-2022-neurons} and internal representations may be biased towards English \citep{wendler,trinley-etal,schut2025do,saji-etal-2025-romanlens} or other languages overrepresented in training data \citep{zhong-etal-2025-language}.

In addition to representational biases, there is evidence that LLMs have English behavioural biases. Representation in multilingual pretraining data has long been observed to be a key factor in determining downstream performance in a language \citep{wu-dredze-2020-languages}. Some multilingual models display English word order preferences even in free word order languages \citep{papadimitriou-etal-2023-multilingual}, and overuse English-like vocabulary and syntax in non-English generated text \citep{guo}.
We build on this line of work by investigating English bias across a range of controlled sentence-level preferences, asking whether models transfer English strategies to other languages.

\subsection{Construction Grammar and LLMs}

Construction Grammar (CxG) models linguistic knowledge as form--meaning pairings \citep{fillmore1988regularity,Goldberg_1995,croft2001radical,goldberg2006constructions}. Recent work has investigated whether LLMs represent English constructions \citep{zhou-etal-2024,scivetti-etal-2026-language,oba-sugawara-2026-cxmp}. However, there are comparatively few works investigating non-English constructions \citep{tseng-etal-2022-cxlm,bunzeck-etal-2025-construction,huang-etal-2025-assessing}, and comparisons across languages are underexplored.
MultiBLiMP \citep{jumelet-etal-2026-multiblimp} evaluates linguistic knowledge across 101 languages, but focuses on the subject--verb agreement strategy across all languages that use it, rather than contrasting different strategies used to express the same meaning across languages.

Recent work has attempted to unite typologically grounded approaches to construction grammar \citep{croft2022morphosyntax} with computational research by creating databases of typologically motivated comparative concepts \citep{lorenzi-etal-2024-mocca}, and investigating annotation of constructions across languages in Universal Dependencies (\citealp{weissweiler-etal-2024-ucxn,croft-nivre-2025-verbal,nivre-2025-constructions}, \textit{i.a.}). We build on these comparative concepts to test whether multilingual LLMs prefer native strategies or overgeneralize strategies from English.

\section{Dataset Creation}
\label{sec:problem-formulation}

We build a controlled dataset for testing whether multilingual language models exhibit cross-lingual bias in the \textbf{strategies} they use to express the same meaning. We focus specifically on transfer involving English: when two languages conventionally use different strategies to express comparable semantic content, does a model distinguish their language-specific preferences, or does it systematically favor the strategy associated with English?

We define the strategy contrasts underlying our evaluation (\S\ref{subsec:constructional-strategies}), then describe how we discover candidate contrasts from learner data (\S\ref{subsec:construction-discovery}) and how the candidates are instantiated and validated by native speakers to construct \dataset{} (\S\ref{subsec:brokenarm}).

\subsection{Constructional Strategies and Cross-Lingual Transfer}
\label{subsec:constructional-strategies}

\begin{table*}[h]
\small
\centering
\setlength{\aboverulesep}{0.2ex}
\setlength{\belowrulesep}{0.2ex}
\begin{tabular}{lll}
\toprule
\textbf{Strategy Source Language} & \textbf{Target language} & \textbf{English} \\
\midrule
German & \cellcolor{lightgreen}Das Wasser fühlt sich kalt an.&\cellcolor{lightred}The water feels itself cold.\\
English&\cellcolor{lightred}Das Wasser fühlt kalt.&\cellcolor{lightgreen}The water feels cold.\\
\midrule
Swedish & \cellcolor{lightgreen}Jag bäddade sängen.&\cellcolor{lightred}I bedded the bed.\\
English&\cellcolor{lightred}Jag gjorde sängen.&\cellcolor{lightgreen}I made the bed.\\
\midrule
French & \cellcolor{lightgreen}Je joue du violoncelle.&\cellcolor{lightred}I play of the cello.\\
English&\cellcolor{lightred}Je joue le violoncelle.&\cellcolor{lightgreen}I play the cello.\\
\midrule
Portuguese & \cellcolor{lightgreen}Está dando caimbra na nossa perna.&\cellcolor{lightred}It is giving cramp in our leg.\\
English&\cellcolor{lightred}Nós temos caimbra nas nossas pernas.&\cellcolor{lightgreen}We have cramps in our legs.\\
\midrule
Chinese & \cellcolor{lightgreen}\zh{那茶喝起来有股烟味。}&\cellcolor{lightred}That tea, when drunk, has a cigarette smell.\\
English&\cellcolor{lightred}\zh{那茶喝起来像香烟。}&\cellcolor{lightgreen}That tea tastes like cigarettes.\\
\midrule
Japanese & \cellcolor{lightgreen}\ja{ハエが目に入っている。}&\cellcolor{lightred}A fly is entering eye.\\
English&\cellcolor{lightred}\ja{私は私の眼の中にハエを持っている。}&\cellcolor{lightgreen}I have a fly in my eye.\\
\bottomrule
\end{tabular}
\caption{
Examples of sentence pairs in each language.
}
\label{tab:constructional-contrasts}
\end{table*}
We consider cases in which two languages use different constructional strategies to express the same semantic content. %and information packaging. 
For each such contrast, we construct four realizations: (i) the native strategy of language $A$ realized in language $A$, (ii) the strategy of language $B$ transferred into language $A$, (iii) the native strategy of language $B$ realized in language $B$, and (iv) the strategy of language $A$ transferred into language $B$. Examples for different language pairs are shown in Table~\ref{tab:constructional-contrasts}.

We define a \textbf{cross-linguistic strategy contrast} as a pair of sentences from two languages using the same strategy, where the native strategy of one language, when transferred to the other language, makes the sentence less acceptable. For example, in Fig.~\ref{fig:page1}, the transitive strategy used in English for instrument playing, when transferred to French, creates the unacceptable sentence \textit{*Je joue le violoncelle.}

Importantly, transferred strategies need not be categorically unacceptable. They may range from acceptable but dispreferred alternatives to strongly unacceptable realizations.

\subsection{Strategy Discovery from Learner Data}
\label{subsec:construction-discovery}

To form an inventory of constructions that are likely to be realized by contrasting strategies across our target languages, we develop a semi-automatic discovery pipeline using the NAIST Lang-8 Learner Corpora \citep{mizumoto-etal-2011-mining}. We compare productions by learners with English as their reported L1 against their native corrections, using recurrent differences in function-word or syntactic realization as candidate strategy contrasts. Individual learner errors are used only for discovery and are not assumed to reflect L1 transfer.

\paragraph{Correction reconstruction.}
As Lang-8 Learner Corpora contain inconsistent correction tags, we reconstruct the corrected sentences using Llama3\textsubscript{70B} Instruct \citep{grattafiori2024llama3herdmodels} and GPT-5.1 \citep{singh2026openaigpt5card} before further filtering.

\paragraph{Automatic candidate classification.}
We parse each learner--correction pair with Stanza \citep{qi-etal-2020-stanza}, word-align by exact match and identify candidate strategy contrasts as differences in function-word realization or dependency relations that cannot be attributed solely to word order. We exclude pairs where only content words are edited, thereby retaining differences involving function words, either alone or together with content words. We further restrict the pairs to a Stanza-tokenized edit distance of at most two and group recurring patterns by UPOS and edit-operation signatures.

\paragraph{Candidate selection.}
We prioritize sufficiently frequent recurring patterns for manual inspection, using Tiny-Aya-Global \citep{tiny-aya} performance to prioritize patterns for which the model frequently fails to prefer the native correction. Model performance serves only as a prioritization signal and does not determine final construction selection. From the inspected candidates, we select constructions based on productivity, the presence of a plausible cross-lingual contrast with English, diversity, and interpretability as a strategy contrast. Detailed procedures for candidate filtering, grouping, ranking, and selection are provided in Appendix~\ref{app:candidate-selection}.

\subsection{\dataset{}}
\label{subsec:brokenarm}

\paragraph{Construction inventory}

These constructions fall into three broader groups: (i)~\textbf{body-part constructions}, in which a body part is realized as an affected entity (e.g., \textit{I brushed my teeth}); (ii)~\textbf{everyday-activity constructions}, covering conventionalized everyday activities (e.g., \textit{play the piano}, \textit{wash the dishes}, and \textit{eat dinner}); and (iii)~\textbf{sensory-experience constructions}, covering stimulus- and experiencer-oriented expressions of perception with contrasts in argument structure and intentionality (e.g., \textit{X smells like Y}, \textit{hear}, \textit{listen to}).

\paragraph{Four-way sentence creation}
For each construction, we create instances following the four-way design in Section~\ref{subsec:constructional-strategies}. Starting from a natural English sentence expressing semantic content \(a\), native speakers produce a natural target-language realization, a target-language realization preserving the English strategy, and an English realization preserving the target-language strategy. This bidirectional design allows us to quantify constructional bias in both the target language and English. 
From this point on, we treat the English and target-language pairs as separate instances for evaluation unless otherwise noted.

\paragraph{Native preference annotation}
\dataset{} was annotated by NLP experts who are native speakers of their respective languages. Because transferred constructions may vary gradiently in acceptability, annotators compare the language-internal and transferred realizations using a six-level preference scale (\(\gg, >, =, <, \ll, \neq\)). While the arrows and \(=\) denote preference judgements on semantically matching sentences, \(\neq\) indicates that both sentences are equally acceptable but do not mean the same thing (e.g., \textit{I had lunch \(\neq\) I ate at midday}). Because we evaluate sentence probabilities without context, and not translation, this makes them unsuitable for our purposes.

We retain instances with a native preference of \(\gg\) or \(>\) for the primary benchmark. The magnitude of this preference need not be symmetric across languages. For example, the German contrast in Table~\ref{tab:constructional-contrasts} may show a weaker native preference than its English counterpart. \dataset{} is therefore parallel in semantic content and strategy, but not necessarily in the magnitude of native preference.

\subsection{Japanese Baseline}
To investigate whether models specifically favor English-derived strategies over strategies transferred from other languages, we additionally construct a Japanese control dataset.

In addition to the native Japanese strategy and the English strategy transferred into Japanese, we construct realizations in Japanese using strategies from German, Swedish, French, Portuguese, and Chinese. For each source language, we select semantic contents for which its strategy differs sufficiently from both the English and Japanese strategies. This allows us to compare transfer from English with transfer from other languages while holding the evaluation language, Japanese, constant.

\section{Methodology}

\subsection{Sequence-Level Strategy Preference}
\label{subsec:seq-level-constructional-pref}

We treat strategy preference as a property of the complete realization rather than of the locally contrasting tokens alone. A strategy choice can affect the probability assigned to subsequent material, even when that material is shared across the two realizations. For example, in a German contrast \textit{Ich habe mir den Arm gebrochen} and \textit{Ich habe meinen Arm gebrochen}, the context \textit{Ich habe} does not yet identify the expression as a body-part construction, and the model's preference between the two strategies may therefore be reflected in the probabilities of the subsequent tokens. Sequence-level scoring therefore allows us to capture the model's preference for each strategy realization as a whole.

We quantify this preference using SLOR \citep{pauls-klein-2012-large, https://doi.org/10.1111/cogs.12414}, which normalizes sequence log probability by sequence length and the unigram probabilities of its lexical material:
\[
\operatorname{SLOR}(S)
=
\frac{
\log p(S)-\log p_{\mathrm{uni}}(S)
}{
|S|
}.
\]

\noindent
This normalization is particularly suitable for our setting because alternative strategy realizations can differ both in length and in the baseline frequency of their lexical material. SLOR thus allows us to compare how strongly a model favors each realization beyond these baseline differences.

We define the model's strategy preference as
\[
\Delta_{\mathrm{SLOR}}
=
\operatorname{SLOR}(S_{native})
-
\operatorname{SLOR}(S_{trans}),
\]
\noindent
where \(S_{native}\) is the native strategy realization and \(S_{trans}\) the transferred strategy realization. Positive values indicate a preference for the native strategy, whereas negative values indicate a preference for the transferred strategy.

\begin{figure*}[t]
    \centering
    \small

    \makebox[\textwidth]{%
        \begin{minipage}{0.49\textwidth}
            \centering
            (a) English-centric models
        \end{minipage}
        \hfill
        \begin{minipage}{0.49\textwidth}
            \centering
            (b) Non-English-centric models
        \end{minipage}
    }

    \includegraphics[width=\textwidth]{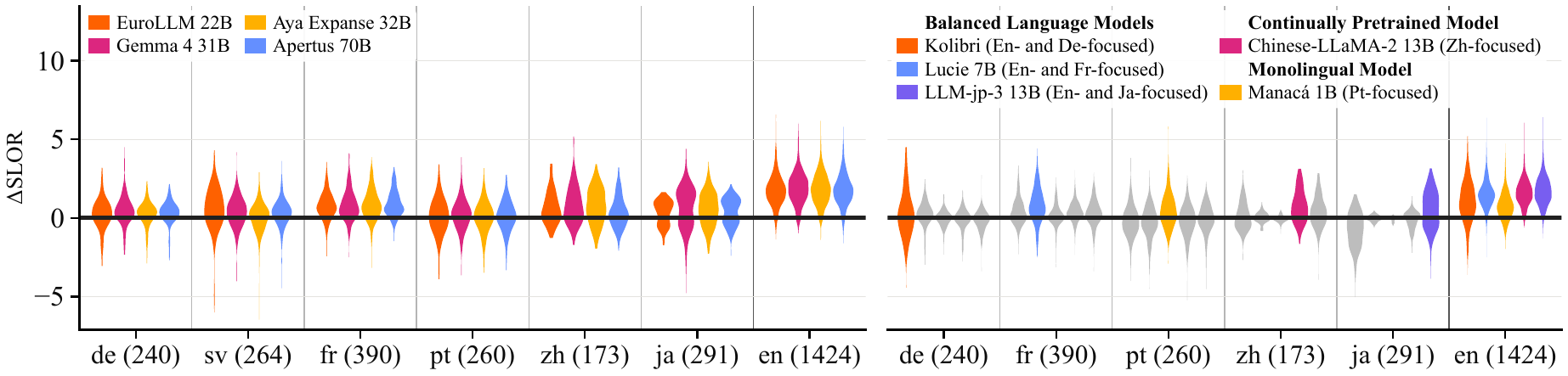}

    \caption{
        Distribution of strategy preferences across languages for
        sentence pairs with strong native-speaker preferences ($\gg$),
        measured by $\Delta$SLOR.
        Following \citet{zhong-etal-2025-language},
	continually pretrained models (CPMs) are adapted
	from English-centric pretrained models through
	additional pretraining on target-language data,
	whereas balanced language models (BLMs) are
	pretrained from scratch with an emphasis on
	English and their respective focus languages.
        Positive values indicate a preference for the native strategy. Numbers in parentheses indicate sample sizes.
    }
    \label{fig:main_violin}
\end{figure*}

\subsection{Factors Associated with Strategy Preferences}

We fit multiple linear regression models to identify factors associated with the models' strategy preferences. Predictors include Model Size (total parameter count, including for MoE models), pretraining language proportion, Instruction Tuning, human preference annotations, and cross-linguistic structural features, including strategy distance, other-language strategy divergence, word-order divergence, and non-1:1 alignment rates.

For pretraining language proportion, we use the base-10 logarithm of language-specific corpus sizes from GlotCC \citep{NEURIPS2024_1e605762} as a proxy for language availability in web-crawled pretraining data. Strategy distance quantifies how different the strategy instantiated in the target language is from native strategies in other languages. Other-language strategy divergence measures the variation among strategies instantiated in other languages to express the same semantic content. Word-order divergence quantifies the proportion of word-order inversions between the native strategies in the target language and English, while the non-1:1 alignment rate measures the proportion of words without one-to-one correspondences.\footnote{Detailed feature definitions, computation procedures, and regression specifications are provided in Appendix \ref{appendix:regression}.}

\section{Results}

\subsection{Cross-Lingual Asymmetries in Strategy Preferences}
\label{subsec:genresults}

\begin{figure*}[h]
    \centering
    \includegraphics[width=\textwidth]{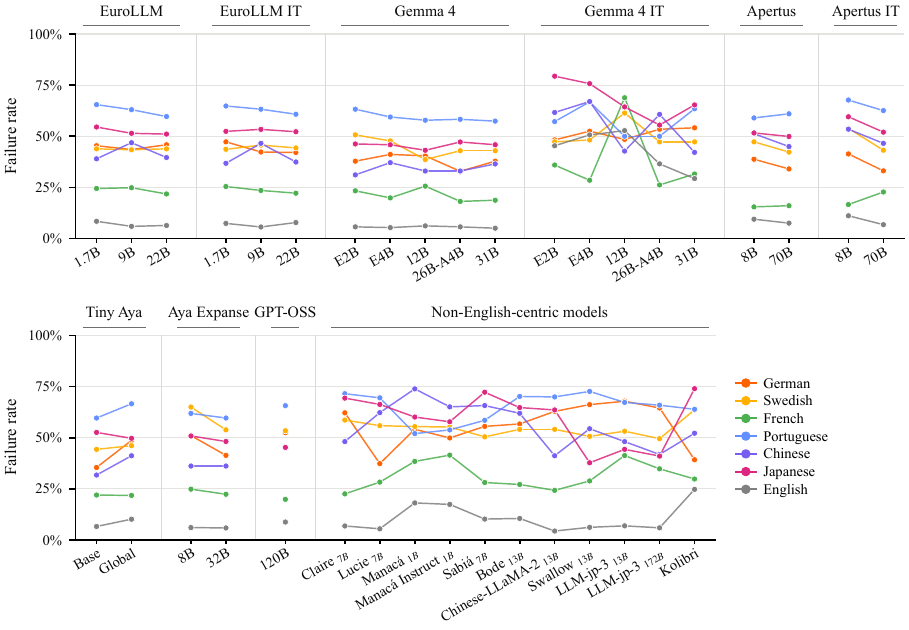}
    \caption{
        Failure rates (\%) across languages, defined as the percentage
        of sentence pairs for which a model assigns a higher SLOR
        to the transferred strategy than to the native strategy.
    }
    \label{fig:failure_rates}
\end{figure*}

We first examine whether language models distinguish native strategies from transferred strategies across languages. We evaluate this ability using two complementary measures: the difference in SLOR between native and transferred strategy realizations (\(\Delta\)SLOR), and the failure rate, defined as the proportion of sentence pairs for which a model assigns a higher SLOR to the transferred strategy than to the native strategy. Fig.~\ref{fig:main_violin} presents the distributions of \(\Delta\)SLOR for the largest model from each model family, while Fig.~\ref{fig:failure_rates} reports failure rates across models and languages.

Fig.~\ref{fig:main_violin}(a) reveals a systematic asymmetry in English-centric models' strategy preferences. While these models generally assign higher SLOR to native English strategies than to strategies transferred from other languages, their preferences are substantially less consistent in non-English languages, including German, Swedish, Portuguese, Japanese, and Chinese. In these languages, the distributions exhibit more frequent negative values, indicating that models often fail to distinguish native strategies from English-transferred alternatives. This asymmetry suggests that English-centric models' ability to distinguish strategies does not appear uniformly across languages. A linear mixed-effects model confirmed significantly higher $\Delta$SLOR in English than the equally weighted mean across non-English languages for English-centric models (Wald $t$-test, $p = 1.09 \times 10^{-46}$, $\mathrm{df} = 227.28$).

However, Fig.~\ref{fig:main_violin}(b) shows that this asymmetry is not universal across model families. Models with a stronger training focus on non-English languages exhibit different preference patterns, often assigning larger positive \(\Delta\)SLOR values in their focus languages than in other languages, although these preferences do not necessarily exceed those observed in English. This contrast suggests that the linguistic composition of model training may be associated with the acquisition of strategy preferences. A linear mixed-effects model confirmed significantly higher $\Delta$SLOR in their focus languages than in other non-English languages (Wald $t$-test, $p = 3.05 \times 10^{-10}$, $\mathrm{df} = 59.09$).

Fig.~\ref{fig:failure_rates} presents the failure rates across models and languages. Both English-centric and non-English-centric models generally exhibit substantially lower failure rates in English than in other languages, indicating that the English advantage persists across model families.

\begin{figure}[h]
    \centering
    \includegraphics[width=\columnwidth]{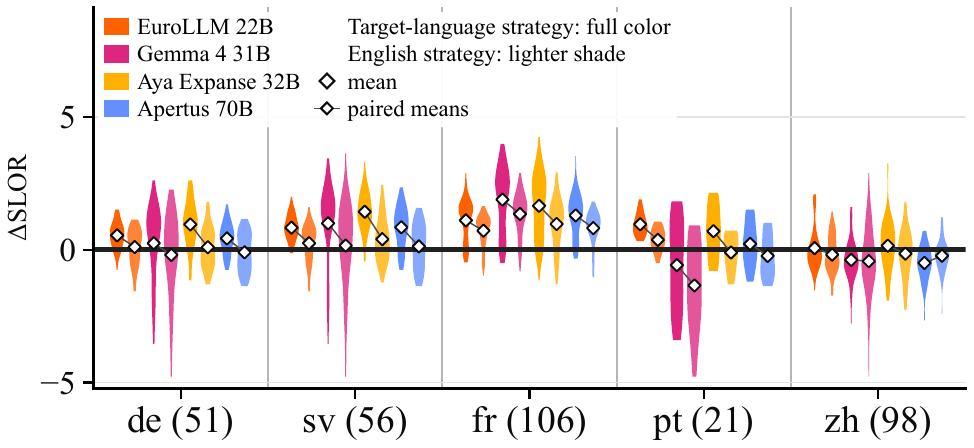}
    \caption{
        Comparison of model preferences for native Japanese strategies
        over strategies transferred from English and other languages,
        measured by $\Delta$SLOR.
        All comparisons are evaluated in Japanese, with matched semantic
        content and restricted to sentence pairs with strong preference on the native Japanese strategy ($\gg$)
        over both the English-transferred and the corresponding
        non-English-transferred strategies.
        Positive values indicate a preference for the native Japanese strategy. Numbers in parentheses indicate sample sizes.
    }
    \label{fig:jacontrol}
\end{figure}

We examine this asymmetry through a controlled comparison in Japanese (Fig. \ref{fig:jacontrol}). Holding the evaluation language, semantic content, and human preferences constant, we compare models' preferences for native Japanese strategies over strategies transferred from other languages. Models often exhibit stronger preferences for native Japanese strategies when compared against transferred strategies from non-English languages than against transferred ones from English (Wald $t$-test, $p = 7.53 \times 10^{-9}$, $\mathrm{df} = 78.84$). This suggests that English-derived strategies are more likely to be favored over native Japanese strategies than transfers from other languages, providing further evidence of English-biased cross-lingual generalization.

\subsection{Human Preferences Are More Strongly Associated with Model Behavior Than Model Scale}
\label{subsec:regression}

\begin{figure}[h]
    \centering
    \includegraphics[width=\columnwidth]{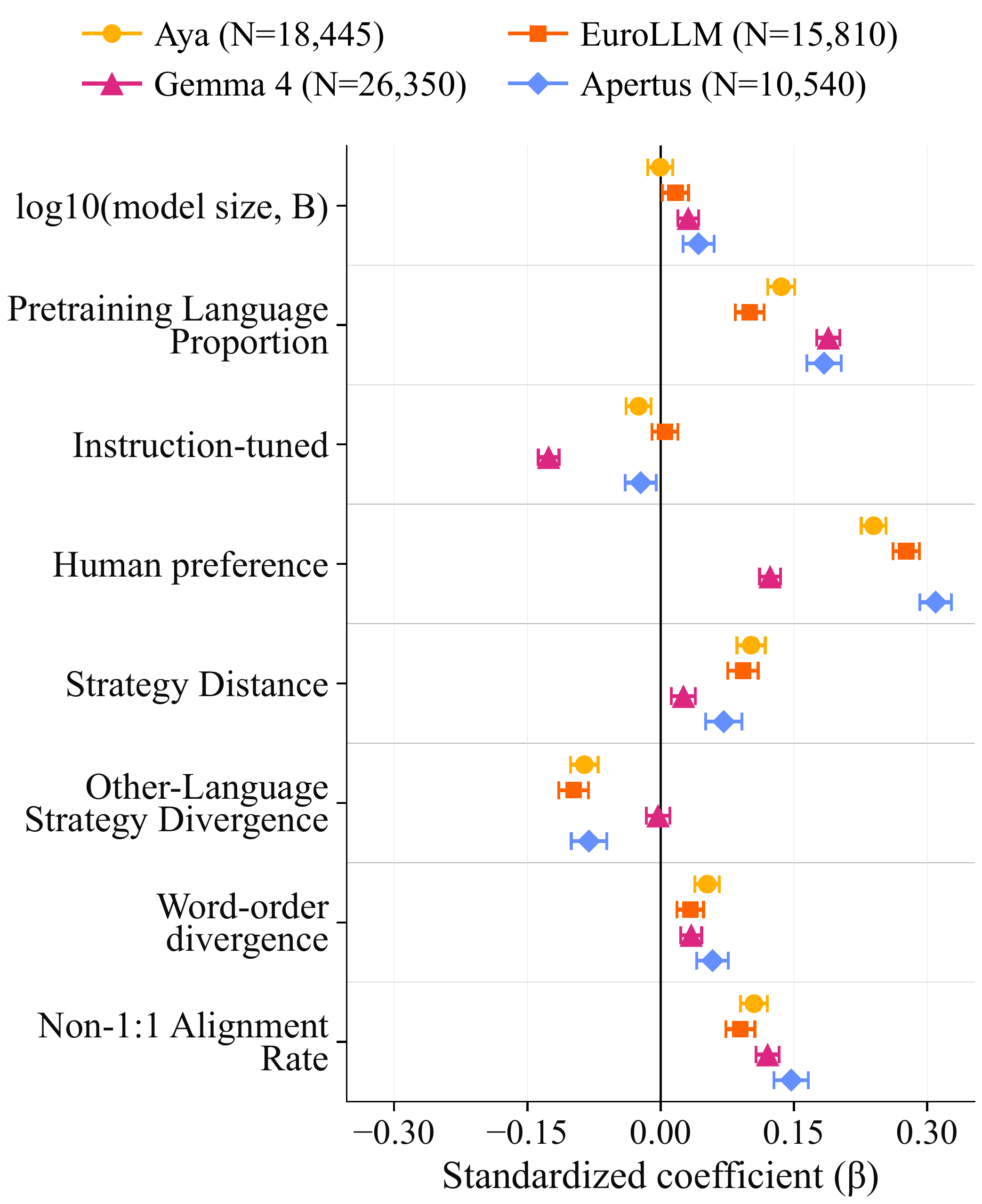}
    \caption{
        Standardized regression coefficients from multivariable linear
        regression models predicting $\Delta$SLOR.
        Predictors include human preference annotations, model size,
        instruction tuning, and cross-linguistic structural features.
        Positive coefficients indicate a positive association with
        preference for the native strategy.
    }
    \label{fig:regression}
\end{figure}

To investigate which factors are associated with models' strategy preferences, we conduct a multivariable regression analysis incorporating human preference annotations, model size, instruction tuning, and linguistically motivated features capturing cross-linguistic structural variation.

As shown in Fig. \ref{fig:regression}, human preference emerges as a strong predictor among the examined features, with estimated coefficients of \(\beta = 0.12- 0.31\) ($p < 0.001$ for all four model families;
two-way cluster-robust $t$-tests). Although the magnitude of this association is modest, the positive coefficient suggests that models tend to assign larger preference differences to contrasts for which native speakers express stronger preferences for the native strategy.
In contrast, model size exhibits little association with model preferences. This finding differs from the scaling-related improvements reported in multilingual linguistic evaluations such as MultiBLiMP \citep{jumelet-etal-2026-multiblimp}, suggesting that increasing model capacity does not necessarily lead to corresponding improvements in distinguishing strategies. Similarly, instruction tuning and the examined linguistic features show relatively small associations with model preferences.

\subsection{English-Centric Models Share Similar Preference Patterns}
\label{subsec:spearman}

\begin{figure}[h]
    \centering
    \includegraphics[width=\columnwidth]{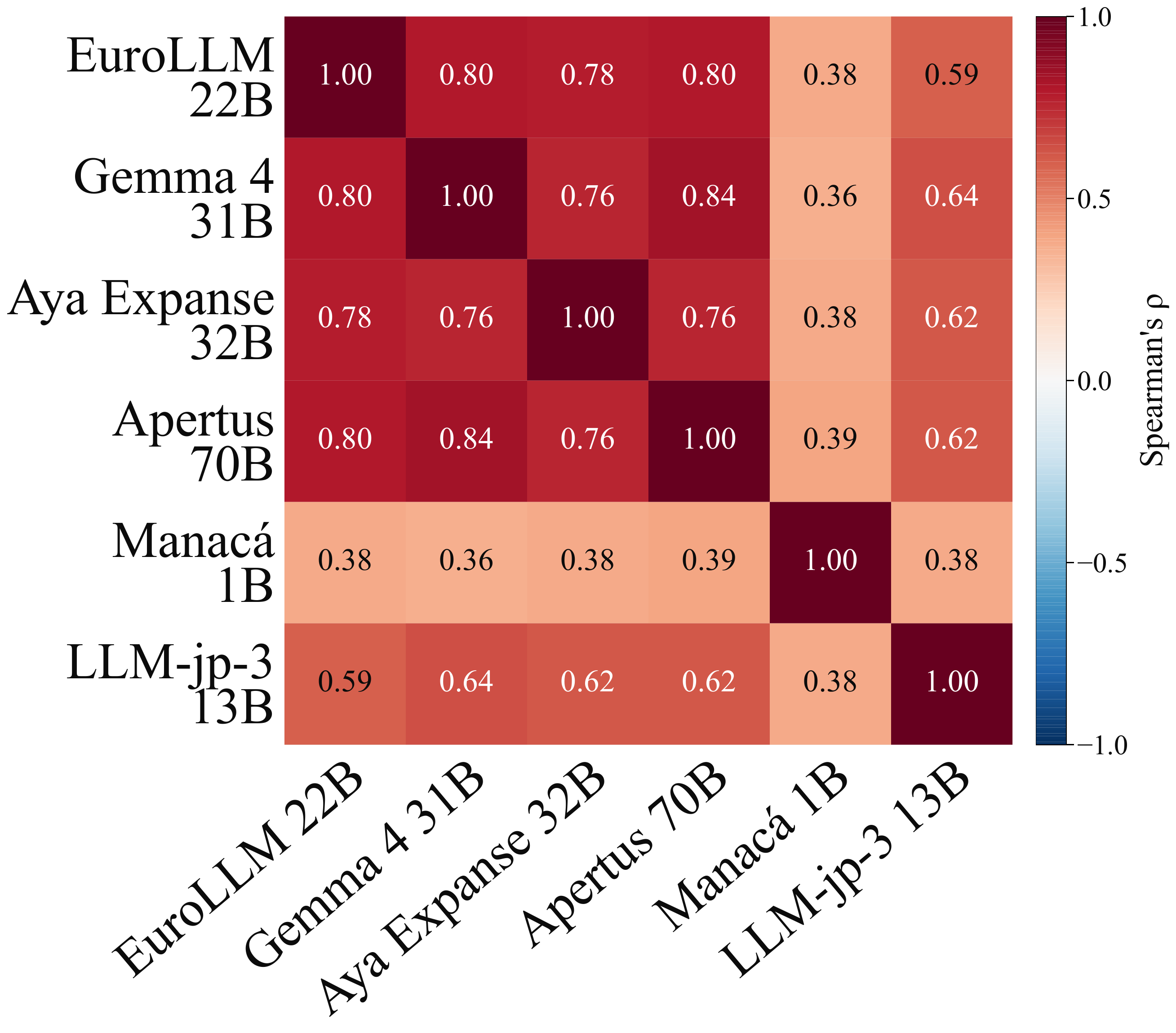}
    \caption{
        Pairwise Spearman rank correlations of $\Delta$SLOR across
        language models, computed over matched sentence pairs.
        Higher correlations indicate greater similarity in strategy preference distributions between models.
    }
    \label{fig:model_correlation}
\end{figure}

We further investigate whether different models exhibit similar preferences across individual strategy contrasts. To this end, we compute pairwise Spearman rank correlations between models' \(\Delta\)SLOR values over matched sentence pairs (Fig. \ref{fig:model_correlation}).

English-centric models exhibit high correlations with one another, with Spearman coefficients ranging from approximately 0.76 to 0.84. These correlations extend across different model families, suggesting that English-centric models share similar relative patterns of preference across strategy contrasts despite differences in architecture and model scale. In other words, sentence pairs that elicit relatively strong native-strategy preferences in one English-centric model tend to elicit similarly strong preferences in other English-centric models.

In contrast, correlations between English-centric and non-English-centric models are somewhat weaker, with coefficients around 0.36 to 0.64. This indicates that models with different linguistic training emphases do not necessarily exhibit the same relative preferences across strategy contrasts.

These findings complement the language-level asymmetries observed in Section \ref{subsec:genresults}. While English-centric models vary in their overall performance, they exhibit substantial similarities in which strategy contrasts are easy or difficult. Lower correlations with non-English-centric models further suggest that these patterns are not universal properties of the evaluated constructions.

Overall, our results reveal systematic asymmetries in multilingual models' strategy preferences. While English-centric models generally favor native strategies more strongly in English than in other languages, non-English-centric models show varying direction and magnitude of these preferences tied to their respective focus languages. Moreover, the substantial correlations across English-centric model families, together with the limited association of model size and instruction tuning with strategy preferences, suggest that shared aspects of training may be more relevant to constructional behavior than model scale. These findings highlight limitations in the robustness of language-specific constructional knowledge acquired under predominantly English-centric multilingual training.

\section{Conclusion}
We introduced \dataset, a new benchmark grounded in the notion of constructional strategies, and evaluated 39 models across seven languages.

Our results reveal systematic constructional bleeding in multilingual LLMs. Models frequently struggle to distinguish native from transferred strategies, with English-centric models exhibiting stronger preferences for native strategies in English than in other languages. Non-English-centric models show different patterns depending on their focus languages. Furthermore, our regression analysis shows that model preferences are more strongly associated with human preferences than with model size, instruction tuning, or other linguistic features. English-centric models also exhibit highly correlated preferences across model families.

Together, these findings suggest that multilingual LLMs acquire strategy knowledge unevenly, often favoring English-associated strategies even when processing non-English languages. The consistency of these patterns across model families highlights limitations of current English-dominated multilingual training practices.

\section*{Limitations}

The set of languages covered in the dataset is limited by the native languages of the paper's authors, and as such, while covering three major typological families, it is not as diverse as would be desirable to find the largest possible set of strategies for each sentence. Likewise, \dataset{} only covers three groups of constructions. Finally, although we report IAA on a subset of our data, our primary annotations were mostly conducted by one native author annotator per language.

\paragraph{Interpretation of results.} Given how the models tested are built, we are not able to report the exact pre-training data proportion for non-English-centric models, as their corpus often includes other languages. Also, we do not disentangle the effects of translationese in the pre-training data from constructional bleeding. Finally, due to time limitations, only one language––Japanese––was used as a fixed language control.

\section*{Ethical Considerations}
Annotation for \dataset{} was carried out primarily by the authors, with one additional contributor (American English native speaker) who voluntarily participated in the annotation process and consented to the use of their annotations for research purposes. For inter-annotator agreement, additional non-author annotators independently annotated a subset of the dataset (50 sentence pairs per language, except for Japanese, which included 100).

No annotators received monetary compensation for the primary annotation. Among the second annotators, one received monetary compensation for annotating a larger amount of Japanese sentence pairs, while the others participated voluntarily.

The non-author annotator spent approximately four hours on primary annotation, while author-annotators spent approximately 8--50 hours each on annotations. Additional non-author annotators for IAA spent approximately 20-30 minutes on the annotation task.

NAIST Lang-8 Learner Corpora were used solely to identify construction types which we focus on, and our dataset does not include sentences extracted from other corpora. Some of the code is written with the help of AI assistants, and verified by the authors.

\section*{Acknowledgments}
The authors acknowledge the financial support by the Federal Ministry of Research, Technology and Space of Germany and by Sächsische Staatsministerium für Wissenschaft, Kultur und Tourismus in the programme Center of Excellence for AI-research ``Center for Scalable Data Analytics and Artificial Intelligence Dresden/Leipzig'', project identification number: \texttt{ScaDS.AI}.
Participation in this research by Georgetown affiliates was supported in part by NSF award \texttt{IIS-2144881}.
Tiago Torrent is a Research Productivity Grantee of the Brazilian National Council for Scientific and Technological Development (CNPq grant \texttt{311241/2025-5}). His participation in this research relies on funding granted to INCT-TILDIAR (CNPq grant \texttt{408490/2024-1}). Marie-Catherine de Marneffe is a Research Associate of the Fonds de la Recherche
Scientifique – FNRS. 
We thank Lori Levin for annotating English preferences and giving us constructive feedback on this paper. We are also thankful to the following colleagues for annotating for the following language preferences used to estimate Inter-Annotator Agreement: Andre Coneglian, for Brazilian Portuguese, Luke Gessler, for English, Norika Hiruta, for Japanese, Linus Högberg, for Swedish, Erika De Longueville, for French, Jacob Lee Suchardt, for German, and Wenhui Zhao, for Chinese.
Part of this work was initiated by Dagstuhl Seminar 25301 ``Linguistics and language models: What can they learn from each other''.

\bibliography{custom,anthology-1,anthology-2,literatur}

\appendix

\section{Appendix}
\label{sec:appendix}

\subsection{Dataset Statistics}
\label{app:dataset_statistics}

Our dataset consists of 669 instances of semantic content, each instantiated across multiple languages and constructional strategies.

Table~\ref{tab:strategy_preferences} reports the number of sentence pairs with strong ($\gg$) and weak ($>$) human preferences for native over transferred strategies, organized by construction group and the languages from which strategies are transferred.

Table~\ref{tab:dataset_composition} summarizes the dataset composition in terms of temporal interpretations, subject types, and construction groups.

\begin{table}[t]
\centering
\small
\setlength{\tabcolsep}{4pt}
\renewcommand{\arraystretch}{1.1}

\begin{tabular}{@{}llrrrrrr@{}}
\toprule
\multicolumn{2}{c}{\textbf{Strategy}} &
\multicolumn{2}{c}{\shortstack{\textbf{Body}\\\textbf{Part}}} &
\multicolumn{2}{c}{\shortstack{\textbf{Everyday}\\\textbf{Activities}}} &
\multicolumn{2}{c}{\shortstack{\textbf{Sensory}\\\textbf{Experience}}} \\
\cmidrule(lr){1-2}
\cmidrule(lr){3-4}
\cmidrule(lr){5-6}
\cmidrule(lr){7-8}
\textbf{Native} &
\textbf{Transferred} &
$\boldsymbol{\gg}$ & $\boldsymbol{>}$ &
$\boldsymbol{\gg}$ & $\boldsymbol{>}$ &
$\boldsymbol{\gg}$ & $\boldsymbol{>}$ \\
\midrule

\multirow{6}{*}{en}
 & de & 162 & 7 & 4 & 36 & 66 & 1 \\
 & se & 78 & 13 & 39 & 22 & 70 & 1 \\
 & fr & 188 & 3 & 182 & 1 & 39 & 14 \\
 & pt & 68 & 4 & 18 & 48 & 85 & 6 \\
 & zh & 60 & 5 & 122 & 7 & 15 & 11 \\
 & ja & 117 & 6 & 65 & 43 & 46 & 8 \\

\midrule
de & \multirow{6}{*}{en} & 29 & 171 & 138 & 0 & 73 & 12 \\
se &  & 65 & 133 & 153 & 9 & 46 & 34 \\
fr & & 69 & 126 & 215 & 7 & 106 & 0 \\
pt & & 31 & 128 & 114 & 40 & 115 & 18 \\
zh & & 34 & 57 & 107 & 65 & 32 & 25 \\
ja & & 54 & 140 & 177 & 45 & 60 & 43 \\

\midrule
\multirow{5}{*}{ja}
 & de & 178 & 0 & 0 & 43 & 30 & 0 \\
 & se & 105 & 74 & 9 & 0 & 20 & 38 \\
 & fr & 179 & 0 & 69 & 51 & 24 & 12 \\
 & pt & 22 & 116 & 9 & 9 & 6 & 36 \\
 & zh & 34 & 16 & 84 & 0 & 26 & 62 \\

\bottomrule
\end{tabular}

\caption{
Distribution of human preferences for native over
transferred strategies across three
construction groups.
$\gg$ and $>$ indicate strong and weak preferences
for the native strategy, respectively.
Japanese-control comparisons use Japanese as the
native strategy and a non-English language as the
source of the transferred strategy.
}
\label{tab:strategy_preferences}
\end{table}

\begin{table}[t]
\centering
\small
\setlength{\tabcolsep}{4pt}
\renewcommand{\arraystretch}{1.1}

\begin{tabular}{@{}lp{0.48\columnwidth}r@{}}
\toprule
\textbf{Category} &
\textbf{Value} &
\textbf{Count} \\
\midrule

\multirow{6}{*}{Temporal}
 & Past & 236 \\
 & Ongoing & 133 \\
 & Recurrent & 121 \\
 & Current state & 77 \\
 & Anterior & 54 \\
 & Generic & 48 \\

\midrule
\multirow{4}{*}{Subject}
 & 1sg pronoun & 195 \\
 & 3sg pronoun & 192 \\
 & 1pl pronoun & 192 \\
 & Common noun & 90 \\

\midrule
\multirow{3}{*}{Group}
 & Everyday activities & 279 \\
 & Body part & 200 \\
 & Sensory experience & 190 \\

\midrule
\multicolumn{2}{l}{\textbf{Total per language}} &
\textbf{669} \\

\bottomrule
\end{tabular}

\caption{
Dataset composition by temporal interpretation,
subject type, and construction group.
The distribution is identical across all evaluated
languages.
}
\label{tab:dataset_composition}
\end{table}

\subsection{List of models}
Table \ref{tab:model_citations} reports the models used in our experiments.

\begin{table*}
\centering
\small
\begin{tabular}{l l l}
\toprule
\textbf{Family} & \textbf{Model} & \textbf{Citation} \\
\midrule
\multirow{2}{*}{Aya Expanse}
  & \texttt{CohereLabs/aya-expanse-8b}  & \multirow{2}{*}{\citet{aya-expanse}} \\
  & \texttt{CohereLabs/aya-expanse-32b} & \\
\midrule
\multirow{5}{*}{Tiny Aya}
  & \texttt{CohereLabs/tiny-aya-base}   & \multirow{5}{*}{\citet{tiny-aya}} \\
  & \texttt{CohereLabs/tiny-aya-global} & \\
  & \texttt{CohereLabs/tiny-aya-earth}  & \\
  & \texttt{CohereLabs/tiny-aya-fire}   & \\
  & \texttt{CohereLabs/tiny-aya-water}  & \\
\midrule
\multirow{6}{*}{EuroLLM}
  & \texttt{utter-project/EuroLLM-1.7B}              & \multirow{6}{*}{\citet{martins2024eurollmmultilinguallanguagemodels}} \\
  & \texttt{utter-project/EuroLLM-1.7B-Instruct}     & \\
  & \texttt{utter-project/EuroLLM-9B}                & \\
  & \texttt{utter-project/EuroLLM-9B-Instruct}       & \\
  & \texttt{utter-project/EuroLLM-22B-2512}          & \\
  & \texttt{utter-project/EuroLLM-22B-Instruct-2512} & \\
\midrule
\multirow{10}{*}{Gemma 4}
  & \texttt{google/gemma-4-E2B}        & \multirow{10}{*}{\citet{gemma4}} \\
  & \texttt{google/gemma-4-E2B-it}     & \\
  & \texttt{google/gemma-4-E4B}        & \\
  & \texttt{google/gemma-4-E4B-it}     & \\
  & \texttt{google/gemma-4-12B}        & \\
  & \texttt{google/gemma-4-12B-it}     & \\
  & \texttt{google/gemma-4-26B-A4B}    & \\
  & \texttt{google/gemma-4-26B-A4B-it} & \\
  & \texttt{google/gemma-4-31B}        & \\
  & \texttt{google/gemma-4-31B-it}     & \\
\midrule
\multirow{4}{*}{Apertus}
  & \texttt{swiss-ai/Apertus-8B-2509}
  & \multirow{4}{*}{\citet{hernandez-cano-etal-2026-apertus}} \\
  & \texttt{swiss-ai/Apertus-8B-Instruct-2509} & \\
  & \texttt{swiss-ai/Apertus-70B-2509} & \\
  & \texttt{swiss-ai/Apertus-70B-Instruct-2509} & \\
\midrule
  & \texttt{openai/gpt-oss-120b}
  & \citet{openai2025gptoss120bgptoss20bmodel} \\
  & \texttt{Aleph-Alpha/Kolibri-1-BF16}
  & \citet{alpha2026kolibrisovereigneuropeanmodel} \\
  & \texttt{hfl/chinese-llama-2-13b}                   & \citet{cui2024efficienteffectivetextencoding} \\
  & \texttt{maritaca-ai/sabia-7b}                      & \citet{10.1007/978-3-031-45392-2_15} \\
  & \texttt{recogna-nlp/bode-13b-alpaca-pt-br-no-peft} & \citet{bode}\\
  & \texttt{OpenLLM-France/Claire-Mistral-7B-0.1}      & \citet{louradour-etal-2024-claire}\\
  & \texttt{OpenLLM-France/Lucie-7B}                   & \citet{gouvert2025lucie7bllmlucietraining} \\
  & \texttt{tokyotech-llm/Swallow-13b-hf}              & \citet{Fujii:COLM2024,Okazaki:COLM2024}\\
  & \texttt{llm-jp/llm-jp-3-13b}                       & \citet{llmjp2024llmjpcrossorganizationalprojectresearch, kiyomaru2024llmjp3} \\              
  & \texttt{llm-jp/llm-jp-3-172b} &\citet{llmjp2024llmjpcrossorganizationalprojectresearch, kiyomaru2024llmjp3172b} \\
  & \texttt{AIInstituteLNCC/manaca-1b-base}            & \citet{manaca}\\
  & \texttt{AIInstituteLNCC/manaca-1b-instruct}
  & \citet{manaca, menezes2026manacainstruct} \\
\bottomrule
\end{tabular}
\caption{Models evaluated, grouped by model family.}
\label{tab:model_citations}
\end{table*}

\subsection{Annotation Scheme}
\label{app:annotation_scheme}

We followed a set of annotation guidelines to ensure consistency and comparability across languages.

\paragraph{Annotation Guidelines}
Annotators were instructed to preserve the natural word order of each language as much as possible, including when applying strategies transferred from other languages. Differences outside the target construction were minimized to keep the sentence pairs as comparable as possible.

For constructions involving determiners, annotators were instructed not to introduce determiners within the target construction in languages that lack them, while preserving natural determiner usage elsewhere. Grammatical number was kept consistent within each pair unless directly relevant to the construction or necessary to preserve the naturalness or the intended meaning.

\paragraph{Preference Annotation}
We used a six-category preference annotation scheme:

\begin{itemize}
    \item $\gg$: The native strategy is strongly preferred.
    \item $>$: The native strategy is weakly preferred.
    \item $=$: Both strategies are equally natural, or their naturalness depends on the intended nuance.
    \item $<$: The transferred strategy is weakly preferred.
    \item $\ll$: The transferred strategy is strongly preferred.
    \item $\neq$: Both sentences are natural but differ in meaning, preventing a direct preference comparison.
\end{itemize}

Since native strategies are defined as the conventionally preferred realizations, annotations of $<$ or $\ll$ generally indicated that the sentence pair required revision. In such cases, annotators typically modified the sentences to ensure that the intended native and transferred strategies were correctly represented. Annotation guidelines were discussed and refined through regular meetings among contributors. To improve annotation efficiency, annotators were allowed to leave sentence pairs marked as "SELECT" when neither $\gg$ nor $>$ applied.

\paragraph{Dataset Expansion and Validation.}
We initially annotated 58 instances of semantic content from scratch, covering the construction groups, temporal interpretations, and subject types. Each instance was realized in four sentences across six languages, resulting in 1,392 realizations.

We subsequently expanded the dataset to 669 instances of semantic content, yielding 16,056 realizations. To facilitate this expansion, we provided GPT-5.1 \citep{singh2026openaigpt5card} with previously annotated examples and used it to generate native and transferred strategy realizations. This approach reduced the need to write every sentence from scratch.

However, GPT-5.1 frequently failed to generate transferred strategies accurately. We therefore manually corrected many of the generated sentences through cross-linguistic comparison before submitting them to native speakers for final correction and validation.

Each language was annotated by at least one native speaker with expertise in NLP. We additionally conducted cross-linguistic checks to improve annotation consistency.

\paragraph{Japanese Control}
For the Japanese-control dataset, sentences using strategies transferred from languages other than English were written manually from scratch.

The Japanese-control annotations were conducted by a native Japanese speaker with at least B2-level proficiency in all evaluated languages except Swedish.

\subsection{Robustness Analysis}
\label{app:robustness}

\begin{figure*}[t]
    \centering
    \small

    \makebox[\textwidth]{%
        \begin{minipage}{0.49\textwidth}
            \centering
            (a) English-centric models
        \end{minipage}
        \hfill
        \begin{minipage}{0.49\textwidth}
            \centering
            (b) Non-English-centric models
        \end{minipage}
    }

    \includegraphics[width=\textwidth]{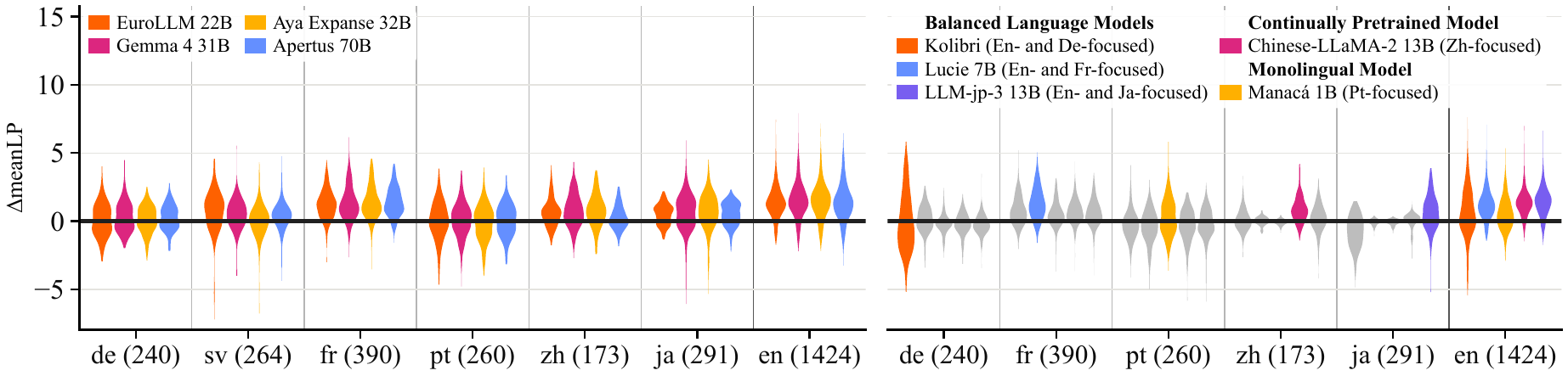}

    \caption{
        Distribution of strategy preferences across languages for
        sentence pairs with strong native-speaker preferences ($\gg$),
        measured by $\Delta$MeanLP.
        Positive values indicate a preference for the native strategy.
    }
    \label{fig:app_violin}
\end{figure*}

To assess the robustness of our findings to the choice of scoring metric, we additionally evaluate model preferences using $\Delta$MeanLP \citep{https://doi.org/10.1111/cogs.12414}.

Fig.~\ref{fig:app_violin} shows a similar asymmetry in strategy preferences to that observed with $\Delta$SLOR in Fig.~\ref{fig:main_violin}. English-centric models generally exhibit stronger preferences for native over transferred strategies in English than in other evaluated languages.

\begin{figure}[h]
    \centering
    \includegraphics[width=\columnwidth]{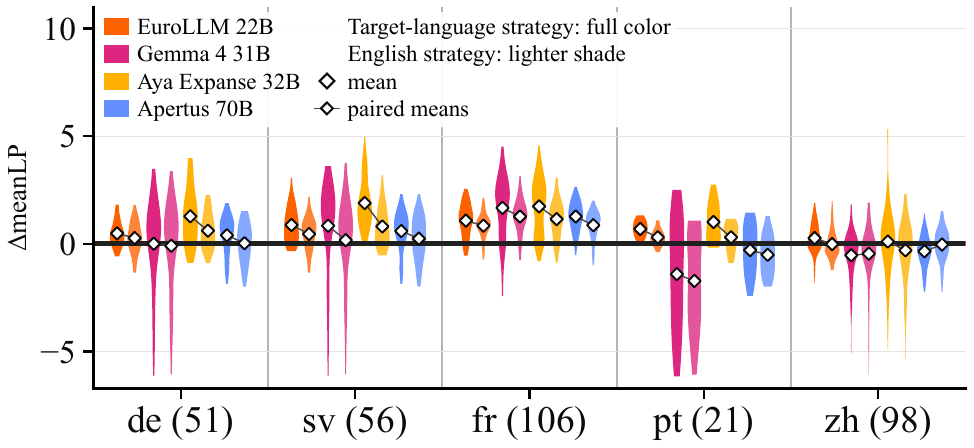}
    \caption{
        Comparison of model preferences for native Japanese strategies
        over strategies transferred from English and other languages,
        measured by $\Delta$MeanLP.
        All comparisons are evaluated in Japanese, with matched semantic
        content and restricted to sentence pairs with a strong preference for the native Japanese strategy ($\gg$)
        over both the English-transferred and the corresponding
        non-English-transferred strategies.
        Positive values indicate a preference for the native Japanese strategy.
    }
    \label{fig:jacontrol-meanLP}
\end{figure}

Fig.~\ref{fig:jacontrol-meanLP} demonstrates that models struggle to distinguish the Japanese native strategies from the English-transferred strategies under $\Delta$MeanLP.

\subsection{Candidate Selection Details}
\label{app:candidate-selection}

\paragraph{Initial filtering.}
We first restrict the candidate pairs to those with an edit distance of at least two and whose constructional differences (defined as defference in dependency relations or function words) involve either only function words or a combination of function and content words. After filtering, the candidate sets contain 4,879 pairs for German, 5,674 for French, 9,632 for Japanese, 551 for Portuguese, 575 for Swedish, and 8,429 for Chinese.

We exclude candidates whose differences involve only content words. Manual inspection showed that such cases yielded relatively few useful strategy contrasts and were predominantly attributable to spelling errors or erroneous linguistic analyses produced by Stanza. Pairs with an edit distance larger than 2 are excluded, as pairs with a larger edit distance tend to contain multiple errors and are therefore more difficult to interpret as a single strategy contrast.

\paragraph{Grouping and ranking.}
We group the remaining candidate pairs according to their operational error patterns, defined by an edit operation and the Part of Speech of the edited word. This results in 157 groups for German, 148 for French, 181 for Japanese, 102 for Portuguese, 116 for Swedish, and 188 for Chinese.

Within each language, we rank groups in ascending order of their average win rate under Tiny-Aya-Global. We define a pair as a win when the model assigns a higher $\Delta_{\mathrm{MeanLP}}$ to the native correction than to the corresponding learner sentence. Low average win rates therefore identify error patterns for which the model relatively often prefers the learner realization over the native correction, making them particularly relevant candidates for identifying strategy preferences that models might struggle in general.

To prioritize sufficiently frequent patterns, we define a language-specific frequency threshold as

$$
\tau_{\ell}
=
\frac{N_{\ell}}{G_{\ell}},
$$

where \(N_{\ell}\) is the total number of candidate pairs and \(G_{\ell}\) is the number of groups for language \(\ell\). The resulting thresholds are shown in Table~\ref{tab:candidate-selection-stats}. Among groups whose frequency exceeds this threshold, we manually inspect the five groups with the lowest win rates for each language.

\begin{table}[h]
\centering
\small
\begin{tabular}{lrrr}
\toprule
\textbf{Language} & \textbf{Candidate pairs} & \textbf{Groups} & \textbf{Threshold} \\
\midrule
German     & 4,879 & 157 & 31.07 \\
French     & 5,674 & 148 & 38.34 \\
Japanese   & 9,632 & 181 & 53.22 \\
Portuguese &   551 & 102 &  5.40 \\
Swedish    &   575 & 116 &  4.96 \\
Chinese    & 8,429 & 188 & 44.84 \\
\bottomrule
\end{tabular}
\caption{Candidate statistics after filtering. The threshold corresponds to the mean number of candidate pairs per group within each language.}
\label{tab:candidate-selection-stats}
\end{table}

\paragraph{Subgroup inspection.}
For each group, we further define subgroups according to the surface forms involved in the corresponding edit operation. We manually inspect all subgroups whose win rate is below 100\%, thereby excluding patterns for which Tiny-Aya-Global consistently prefers the native correction.

From these candidates, we select constructions according to four criteria: (i)~\textit{productivity}, such that the construction can be instantiated with multiple lexical items and contexts; (ii)~\textit{cross-lingual contrast}, such that the relative preference between the two strategies can plausibly reverse in English; (iii)~\textit{diversity}, to avoid selecting multiple near-duplicate phenomena; and (iv)~\textit{interpretability}, such that the contrast admits a clear constructional analysis.

\paragraph{Construction inventory.}
This procedure yields 51 candidate constructions across the six target languages. We instantiate these constructions in 105 sentences and annotate the resulting four-way strategy contrasts and sentence-level preferences. We then examine the inventory cross-linguistically and organize the constructions that are most relevant to our research questions into three broader construction groups:

\begin{enumerate}
\item \textbf{Body part as patient.}
Constructions involving body parts as affected entities, including expressions such as \textit{I broke my arm} and \textit{I brushed my teeth}.

\item \textbf{Everyday activities with salient patients.}
Constructions describing common activities involving a salient, often generic or superordinate patient, such as \textit{play the piano}, \textit{play soccer}, \textit{wash the dishes}, and \textit{eat dinner}. The patient may also have a specific referent in appropriate contexts (e.g., \textit{play the piano you gave me}).

\item \textbf{Sensory experience.}
Constructions expressing sensory perception. These include stimulus-subject structures, optionally with an experiencer (e.g., \textit{something looks/smells/tastes/sounds/feels like X}), and experiencer-subject structures involving verbs such as \textit{see}, \textit{look at}, \textit{watch}, \textit{hear}, and \textit{listen to}. Some English verbs participate in multiple frames, and some contrasts additionally encode differences in intentionality, such as \textit{hear} vs.\ \textit{listen to} and \textit{see} vs.\ \textit{look at}.

\end{enumerate}

Constructions collected during the discovery process that do not fall into these three groups are retained in the released dataset under the category \textsc{Other}.

\subsection{Regression Features}
\label{appendix:regression}

\textbf{Pretraining Language Proportion} uses the sum of the context-length field across records \citep{NEURIPS2024_1e605762}.

\textbf{Strategy Distance} is measured by the mean normalized Levenshtein distance between the corresponding English literal translations of the target-language strategies:
\[
D_l=\frac{1}{|L|-1}\sum_{k\in L,\;k\neq l}
\frac{\operatorname{Lev}(S_l,S_k)}{\max(|S_l|,|S_k|)},
\]
where \(L\) is the set of target languages that meet the strategy preference contrast and \(S_l\) and \(S_k\) denote the corresponding English literal translation for language \(l\) and the other languages \(k\), respectively.

To distinguish language-specific divergence from the structural diversity of the remaining languages, we additionally compute \textbf{Other-Language Strategy Divergence}:

\begin{equation}
\begin{aligned}
\mathrm{OLSD}_l
&= \sum_{\substack{k<m\\k,m\in L\setminus\{l\}}}
\frac{\operatorname{Lev}(S_k,S_m)}
{\max(|S_k|,|S_m|)}\\
&\quad \times \frac{2}{(|L|-1)(|L|-2)}.
\end{aligned}
\end{equation}

Finally, using Awesome-Align \citep{dou-neubig-2021-word} between the English and target-language native-strategy realizations (the tokenization was performed by Stanza), we compute \textbf{Word-Order Divergence},
\[
WO=\frac{\#\text{ inverted aligned token pairs}}{\binom{n_{\mathrm{aligned}}}{2}},
\]
where \(n_{\mathrm{aligned}}\) is the number of aligned token pairs, and \textbf{Non-1:1 Alignment Rate},
\[
A=\frac{\#\text{ non-1:1 target-language tokens}}
{\#\text{ target-language tokens}}.
\]

Multivariable regression was performed only on sentence pairs with complete metadata for all those predictors. Together, these predictors allow us to examine whether strategy preferences are associated with model-level properties, or structural properties of the cross-linguistic contrast.

\subsection{Inter-Annotator Agreement}

We sampled 50 semantically matched canonical sentence pairs from each target language, comprising 16 body-part constructions, 17 everyday-activity constructions, and 17 sensory-experience constructions. For Japanese, we sampled an additional 50 sentence pairs from the Japanese control split, including 10 contrasts for each of the five strategy-transfer languages. Canonical sentences typically feature first-person singular subjects and past-tense verbs, although some contain third-person common-noun subjects. Sentence pairs were randomly selected from this pool of canonical sentences.

We evaluate inter-annotator agreement (IAA) by distinguishing strong native preference ($\gg$) from all other labels, as well as preferences for the native strategy ($\gg$ or $>$) from all other labels. Table~\ref{tab:iaa} reports percentage agreement and Cohen's $\kappa$.

The results show relatively high observed agreement across languages ($0.74-0.98$) for both types of IAA, while Cohen's $\kappa$ varies more substantially, partly reflecting differences in label distributions across languages in our dataset, which predominantly contains sentences favoring native strategies. In particular, the Japanese control split shows high agreement for both strong preference and preference direction, with substantial Cohen's $\kappa$ for strong preference but Cohen's $\kappa$ = $0$ for preference direction. This stems from our dataset construction: the first annotator selected semantic contents that instantiate strategies that are distinct from Japanese and English, resulting in a split exclusively containing preferences towards native strategies.

\begin{table}[h]
\centering
\small
\begin{tabular}{l r rr rr}
\toprule
& & \multicolumn{2}{c}{\textbf{$\gg$ vs others}}
    & \multicolumn{2}{c}{\textbf{($\gg$ or $>$) vs others}} \\
\cmidrule(lr){3-4}\cmidrule(lr){5-6}
\textbf{Lang.} & \textbf{$N$}
    & \textbf{Agr.} & \textbf{$\kappa$}
    & \textbf{Agr.} & \textbf{$\kappa$} \\
\midrule
en & 50  & 0.80 & 0.567 & 0.78 & 0.556 \\
de & 50  & 0.82 & 0.613 & 0.78 & 0.551 \\
se & 50  & 0.78 & 0.428 & 0.94 & 0.849 \\
fr & 50  & 0.82 & 0.526 & 0.98 & 0.929 \\
pt & 50  & 0.76 & 0.525 & 0.86 & 0.451 \\
zh & 50  & 0.74 & 0.475 & 0.84 & 0.649 \\
ja & 50 & 0.80 & 0.597 & 0.86 & 0.610 \\
ja-control & 50 & 0.86 & 0.703 & 0.70 & 0.000 \\
\bottomrule
\end{tabular}
\caption{Inter-annotator agreement by target language (Lang). Strong preference distinguishes $\gg$ from all other labels, whereas preference direction distinguishes $\gg$ or $>$ from all other labels. Agreement (Agr) denotes the proportion of matching annotations, and $\kappa$ denotes Cohen's kappa.}
\label{tab:iaa}
\end{table}

\subsection{Automatic Glossing}
\label{subsec:automatic-glossing}
Our dataset includes linguistic glosses following the Leipzig Glossing Rules \cite{leipzigglossing}. To reduce variation in automatic glossing, we define explicit rules for segmenting overt morphemes and annotating grammatical properties such as case, tense, aspect, and reflexivity, with additional language-specific conventions. Function words and inflectional morphology receive detailed grammatical annotations, while lexical morphemes are generally glossed with English lexical equivalents. We generate the glosses using GPT-5.1 \citep{singh2026openaigpt5card}.

\subsection{Implementation Details}

\paragraph{SLOR and MeanLP Calculation.}
We load models using Hugging Face Transformers \citep{wolf-etal-2020-transformers} and run them in evaluation mode with BF16 precision on GPUs. Sentence log probabilities are computed by summing next-token log probabilities from causal language models, using each model's tokenizer. We tokenize with \texttt{add\_special\_tokens=True}, excluding the initial token from scoring, and convert logits to FP32 before applying log-softmax. For SLOR normalization, we estimate language- and tokenizer-specific unigram distributions from the FLORES-200 development split \citep{nllbteam2022languageleftbehindscaling}, using \texttt{add\_special\_tokens=False} and add-one smoothing ($\alpha=1$). We use greedy decoding and batch size 1. Periods were excluded during preprocessing of the dataset. We use simplified Chinese characters for Mandarin.

\paragraph{Metadata Construction.}
We tokenize sentences using Stanza and obtain word alignments using Awesome-Align with the \texttt{awesome-without-co} checkpoint, extracting representations from Transformer layer 8.

\paragraph{LMMs and Wald $t$-tests.}
In Section \ref{subsec:genresults}, we fit three linear mixed-effects models (LMMs) using the \texttt{lme4} package \citep{JSSv067i01} in R with restricted maximum likelihood (REML) estimation. We evaluate the corresponding fixed-effect contrasts using Wald $t$-tests.

The first analysis examines whether $\Delta\mathrm{SLOR}$ differs significantly between English and non-English languages in English-centric models. We fit

\[
\begin{aligned}
y_{m\ell i} ={}&
\beta_0+\beta_\ell
+u_m+v_{m\ell}+w_i+\varepsilon_{m\ell i},
\end{aligned}
\]

where $y_{m\ell i}$ denotes $\Delta\mathrm{SLOR}$ for model $m$, language $\ell$, and item $i$. English is the reference level, such that $\beta_{\mathrm{en}}=0$, $\beta_0$ represents the English intercept, and $\beta_\ell$ represents the difference between language $\ell$ and English. The terms $u_m$, $v_{m\ell}$, and $w_i$ denote random intercepts for model, model--language combination, and item, respectively, and $\varepsilon_{m\ell i}$ denotes the residual error. We test

\[
H_0:
\frac{1}{6}
\sum_{\ell\ne\mathrm{en}}\beta_\ell=0.
\]

The second analysis examines whether the focus language of a non-English-centric model exhibits a significant difference in $\Delta\mathrm{SLOR}$. We fit

\[
\begin{aligned}
y_{m\ell i} ={}&
\beta_0+\beta_\ell
+\beta_{\mathrm{focus}}F_{m\ell}\\
&+u_m+v_{m\ell}+w_i+\varepsilon_{m\ell i},
\end{aligned}
\]

where $F_{m\ell}$ is a binary indicator equal to one when language $\ell$ is the focus language of model $m$, and zero otherwise. The random intercepts have the same interpretation as in the first LMM. We test

\[
H_0:\beta_{\mathrm{focus}}=0.
\]

The third analysis examines whether the difference in $\Delta\mathrm{SLOR}$ between the Japanese-strategy and English-strategy conditions differs significantly from the difference between the Japanese-strategy and non-English-strategy conditions. We define

\[
\begin{aligned}
y_{m\ell i} ={}&
\Delta\mathrm{SLOR}^{\mathrm{JA\ control}}_{m\ell i}-\Delta\mathrm{SLOR}^{\mathrm{JA\ baseline}}_{m\ell i},
\end{aligned}
\]

and fit

\[
\begin{aligned}
y_{m\ell i} ={}&
\beta_\ell+u_m+v_{m\ell}+w_i+\varepsilon_{m\ell i}.
\end{aligned}
\]

Unlike the first two LMMs, this model does not include a common fixed intercept, so $\beta_\ell$ represents the language-specific fixed effect. We test

\[
H_0:
\frac{1}{L}\sum_{\ell=1}^{L}\beta_\ell=0,
\]

where $L$ denotes the number of languages.

Degrees of freedom were approximated by Satterthwaite's method \citep{64768ed2-5a9f-38a1-bd23-4e118811f875}, as implemented in the R package \textit{ImerTest} \citep{JSSv082i13}.

\paragraph{Multivariable Regression and
Two-Way Cluster-Robust $t$-Tests.}
In Section \ref{subsec:regression}, we fit separate multivariable linear regression models for each of the four model families (Aya, EuroLLM, Gemma 4, and Apertus) using ordinary least squares (OLS). The regression model is specified as

\[
\begin{aligned}
\Delta\mathrm{SLOR}_{i,f}
={}&\beta_{0,f}\\
&+\beta_{1,f}\log_{10}(\mathrm{Size}_{i,f})\\
&+\beta_{2,f}\mathrm{IT}_{i,f}\\
&+\beta_{3,f}\mathrm{Preference}_{i,f}\\
&+\cdots+\varepsilon_{i,f},
\end{aligned}
\]

where $f$ denotes the model family, $i$ denotes an observation, and $\varepsilon_{i,f}$ is the error term. Regression coefficients are estimated using OLS.

Following \citep{cameron2011robust}, we assess statistical significance using two-way cluster-robust $t$-tests, clustering by model and sentence-pair identity. Specifically, we implement the Cameron--Gelbach--Miller (CGM) two-way cluster-robust covariance estimator to account for dependence within both clustering dimensions.

We test the null hypothesis

\[
H_0:\beta_{3,f}=0.
\]

When computing the two-way cluster-robust covariance matrix, we apply a positive semidefinite (PSD) correction by replacing negative eigenvalues with zero. We use $df=\min(M_f,P_f)-1$, where $M_f$ and $P_f$ denote the numbers of model and sentence-pair clusters, respectively, in family $f$.

\paragraph{Computational Resources.}
Model evaluations were conducted in two computing environments. The first was a server equipped with two NVIDIA RTX PRO 5000 Blackwell GPUs (48 GB VRAM each), a 32-core AMD EPYC CPU, and 64 GB of RAM. The second was an HPC cluster, where jobs were allocated 2-4 NVIDIA H100 GPUs, 16 CPU cores, and 192 GB of RAM. Evaluation times varied across models, ranging from two to over eight hours per model, including model loading and inference.

\begin{table*}[h]
\small
\centering
\setlength{\aboverulesep}{0.2ex}
\setlength{\belowrulesep}{0.2ex}
\begin{tabular}{llcclc}
\toprule
\textbf{Lang.} &
\textbf{Native strategy} &
\textbf{A1} &
\textbf{A2} &
\textbf{Transferred strategy} &
\textbf{Failed} \\
\midrule

\multirow{3}{*}{de}
& \cellcolor{lightgreen}Ich habe Zähne geputzt.
& $>$ & $>$
& \cellcolor{lightred}Ich habe meine Zähne geputzt.
& 27/28 \\
& \cellcolor{lightgreen}Ich bin umgeknickt.
& $\gg$ & $\gg$
& \cellcolor{lightred}Ich habe meinen Knöchel umgeknickt.
& 26/28 \\
& \cellcolor{lightgreen}Ich habe mir das Gesicht gewaschen.
& $>$ & $>$
& \cellcolor{lightred}Ich habe mein Gesicht gewaschen.
& 26/28 \\

\midrule

\multirow{3}{*}{sv}
& \cellcolor{lightgreen}Axeln gick ur led på mig.
& $\gg$ & $>$
& \cellcolor{lightred}Jag dislokerade min axel.
& 28/28 \\
& \cellcolor{lightgreen}Maten luktade vitlök.
& $>$ & $>$
& \cellcolor{lightred}Måltiden luktade som vitlök.
& 27/28 \\
& \cellcolor{lightgreen}Jag borstade tänderna.
& $>$ & $>$
& \cellcolor{lightred}Jag borstade mina tänder.
& 25/28 \\

\midrule

\multirow{3}{*}{fr}
& \cellcolor{lightgreen}J'avais une crampe à la jambe.
& $>$ & $\gg$
& \cellcolor{lightred}J'avais une crampe dans ma jambe.
& 25/28 \\
& \cellcolor{lightgreen}J'avais une mouche dans l'oeil.
& $\gg$ & $\gg$
& \cellcolor{lightred}J'avais une mouche dans mon oeil.
& 22/28 \\
& \cellcolor{lightgreen}Je me suis coupé les cheveux.
& $\gg$ & $\gg$
& \cellcolor{lightred}J'ai coupé mes cheveux.
& 19/28 \\

\midrule

\multirow{3}{*}{pt}
& \cellcolor{lightgreen}A voz dela parecia de criança.
& $>$ & $\gg$
& \cellcolor{lightred}A voz dela soava como a voz de uma criança.
& 28/28 \\
& \cellcolor{lightgreen}Eu lavei o rosto.
& $>$ & $\gg$
& \cellcolor{lightred}Eu lavei meu rosto.
& 28/28 \\
& \cellcolor{lightgreen}Eu desloquei o ombro.
& $>$ & $\gg$
& \cellcolor{lightred}Eu desloquei meu ombro.
& 27/28 \\

\midrule

\multirow{3}{*}{zh}
& \cellcolor{lightgreen}\zh{我挠了背}
& $>$ & $>$
& \cellcolor{lightred}\zh{我挠了我的背}
& 28/28 \\
& \cellcolor{lightgreen}\zh{我刷了牙}
& $>$ & $>$
& \cellcolor{lightred}\zh{我刷了我的牙}
& 27/28 \\
& \cellcolor{lightgreen}\zh{我昨天拉了大提琴。}
& $\gg$ & $>$
& \cellcolor{lightred}\zh{我昨天玩了大提琴。}
& 26/28 \\

\midrule

\multirow{3}{*}{ja}
& \cellcolor{lightgreen}\ja{チェスをした。}
& $>$ & $\gg$
& \cellcolor{lightred}\ja{チェスを遊びました。}
& 28/28 \\
& \cellcolor{lightgreen}\ja{背中を掻いた。}
& $>$ & $>$
& \cellcolor{lightred}\ja{私は私の背中を掻いた。}
& 28/28 \\
& \cellcolor{lightgreen}\ja{髪をとかした。}
& $>$ & $>$
& \cellcolor{lightred}\ja{私は私の髪をとかした。}
& 28/28 \\

\bottomrule
\end{tabular}
\caption{
High-failure-rate canonical sentence examples across 28 English-centric models in six languages. For each language, the three sentence pairs are shown, for which both annotators preferred the native strategy ($>$ or $\gg$). The \textbf{A1} and \textbf{A2} columns report the respective native-speaker preference annotations. The \textbf{Failed} column indicates the number of models assigning a lower SLOR to the native strategy than to the transferred strategy, out of 28 evaluated models.
}
\label{tab:constructional-bleeding-examples}
\end{table*}

\subsection{Failed Examples}
Table \ref{tab:constructional-bleeding-examples} presents high-failure-rate canonical sentence examples across 28 English-centric models, along with preference annotations assigned by two native annotators.

\subsection{Failure Rates}
Table \ref{tab:failure_rates} presents the failure rates across models.

\begin{table*}[h]
\centering
\footnotesize
\setlength{\tabcolsep}{5pt}
\begin{tabular}{lrcrrrrrrr}
\toprule
\multicolumn{2}{l}{\textbf{Model}} & \textbf{IT} & \textbf{EN} & \textbf{DE} & \textbf{SV} & \textbf{FR} & \textbf{PT} & \textbf{JA} & \textbf{ZH} \\
\midrule
\multicolumn{10}{l}{\textit{Multilingual models}} \\
\midrule
\multirow{5}{*}{Tiny Aya} & \textit{Base} &  & \failcell{6.6} & \failcell{35.5} & \failcell{44.3} & \failcell{22.0} & \failcell{59.6} & \failcell{52.6} & \failcell{31.8} \\
 & \textit{Earth} & \checkmark  & \failcell{9.5} & \failcell{48.2} & \failcell{46.6} & \failcell{22.9} & \failcell{66.8} & \failcell{49.7} & \failcell{40.9} \\
 & \textit{Fire} & \checkmark & \failcell{8.1} & \failcell{44.2} & \failcell{48.9} & \failcell{23.5} & \failcell{67.3} & \failcell{50.9} & \failcell{40.3} \\
 & \textit{Global} & \checkmark & \failcell{10.2} & \failcell{48.7} & \failcell{46.1} & \failcell{21.8} & \failcell{66.6} & \failcell{49.7} & \failcell{41.2} \\
 & \textit{Water} & \checkmark & \failcell{10.0} & \failcell{47.5} & \failcell{47.3} & \failcell{22.2} & \failcell{66.6} & \failcell{53.2} & \failcell{40.3} \\
\midrule
\multirow{2}{*}{Aya Expanse} & 8B &  & \failcell{6.1} & \failcell{51.1} & \failcell{65.0} & \failcell{24.9} & \failcell{61.9} & \failcell{50.9} & \failcell{36.2} \\
 & 32B &  & \failcell{5.9} & \failcell{41.4} & \failcell{53.9} & \failcell{22.4} & \failcell{59.6} & \failcell{48.2} & \failcell{36.2} \\
\midrule
\multirow{10}{*}{Gemma 4} & E2B &  & \failcell{5.7} & \failcell{37.8} & \failcell{50.7} & \failcell{23.3} & \failcell{63.2} & \failcell{46.2} & \failcell{31.1} \\
 & E2B & \checkmark & \failcell{45.4} & \failcell{48.2} & \failcell{47.3} & \failcell{35.9} & \failcell{57.2} & \failcell{79.4} & \failcell{61.6} \\
 & E4B &  & \failcell{5.4} & \failcell{41.1} & \failcell{47.7} & \failcell{19.9} & \failcell{59.4} & \failcell{45.9} & \failcell{37.1} \\
 & E4B & \checkmark & \failcell{50.7} & \failcell{52.5} & \failcell{48.2} & \failcell{28.5} & \failcell{66.8} & \failcell{75.7} & \failcell{67.0} \\
 & 12B &  & \failcell{6.2} & \failcell{40.4} & \failcell{38.6} & \failcell{25.6} & \failcell{57.8} & \failcell{43.2} & \failcell{33.0} \\
 & 12B & \checkmark & \failcell{52.8} & \failcell{48.5} & \failcell{61.4} & \failcell{68.8} & \failcell{50.0} & \failcell{64.4} & \failcell{42.8} \\
 & 26B-A4B &  & \failcell{5.7} & \failcell{32.9} & \failcell{43.0} & \failcell{18.2} & \failcell{58.3} & \failcell{47.2} & \failcell{33.0} \\
 & 26B-A4B & \checkmark & \failcell{36.5} & \failcell{53.4} & \failcell{47.3} & \failcell{26.2} & \failcell{50.0} & \failcell{55.5} & \failcell{60.7} \\
 & 31B &  & \failcell{5.1} & \failcell{37.8} & \failcell{43.0} & \failcell{18.7} & \failcell{57.4} & \failcell{45.9} & \failcell{36.5} \\
 & 31B & \checkmark & \failcell{29.3} & \failcell{54.1} & \failcell{47.3} & \failcell{31.5} & \failcell{63.5} & \failcell{65.3} & \failcell{42.1} \\
\midrule
\multirow{4}{*}{Apertus} & 8B &  & \failcell{9.5} & \failcell{38.8} & \failcell{47.3} & \failcell{15.5} & \failcell{59.0} & \failcell{51.6} & \failcell{51.3} \\
 & 8B & \checkmark & \failcell{11.1} & \failcell{41.4} & \failcell{53.9} & \failcell{16.6} & \failcell{67.7} & \failcell{59.5} & \failcell{53.5} \\
 & 70B &  & \failcell{7.5} & \failcell{34.0} & \failcell{42.3} & \failcell{16.1} & \failcell{61.0} & \failcell{49.9} & \failcell{45.0} \\
 & 70B & \checkmark & \failcell{6.8} & \failcell{33.1} & \failcell{43.2} & \failcell{22.8} & \failcell{62.6} & \failcell{52.0} & \failcell{46.5} \\
\midrule
\multirow{6}{*}{EuroLLM} & 1.7B &  & \failcell{8.4} & \failcell{45.4} & \failcell{43.9} & \failcell{24.5} & \failcell{65.5} & \failcell{54.5} & \failcell{39.0} \\
 & 1.7B & \checkmark & \failcell{7.4} & \failcell{47.3} & \failcell{43.6} & \failcell{25.4} & \failcell{64.8} & \failcell{52.4} & \failcell{36.8} \\
 & 9B &  & \failcell{6.0} & \failcell{43.5} & \failcell{43.4} & \failcell{24.9} & \failcell{63.0} & \failcell{51.4} & \failcell{46.9} \\
 & 9B & \checkmark & \failcell{5.7} & \failcell{42.3} & \failcell{45.7} & \failcell{23.5} & \failcell{63.2} & \failcell{53.4} & \failcell{46.5} \\
 & 22B &  & \failcell{6.4} & \failcell{45.9} & \failcell{43.9} & \failcell{21.8} & \failcell{59.6} & \failcell{51.1} & \failcell{39.6} \\
 & 22B & \checkmark & \failcell{7.8} & \failcell{42.1} & \failcell{44.3} & \failcell{22.2} & \failcell{60.8} & \failcell{52.2} & \failcell{37.4} \\
\midrule
GPT-oss & 120B & \checkmark & \failcell{8.8} & \failcell{52.5} & \failcell{53.4} & \failcell{19.9} & \failcell{65.7} & \failcell{45.3} & \failcell{45.3} \\
\midrule
\multicolumn{10}{l}{\textit{Non-English-centric models}} \\
\midrule
Kolibri 1 & 70B & \checkmark & \failcell{24.8} & \failcell{39.2} & \failcell{63.6} & \failcell{29.8} & \failcell{63.9} & \failcell{74.0} & \failcell{52.2} \\
Claire-Mistral & 7B & & \failcell{6.9} & \failcell{62.2} & \failcell{58.6} & \failcell{22.6} & \failcell{71.5} & \failcell{69.4} & \failcell{48.1} \\
Lucie & 7B & & \failcell{5.5} & \failcell{37.4} & \failcell{55.9} & \failcell{28.3} & \failcell{69.5} & \failcell{66.3} & \failcell{62.3} \\
Sabiá & 7B & & \failcell{10.2} & \failcell{55.6} & \failcell{50.5} & \failcell{28.1} & \failcell{58.5} & \failcell{72.3} & \failcell{65.7} \\
Bode & 13B & \checkmark & \failcell{10.5} & \failcell{56.7} & \failcell{54.1} & \failcell{27.2} & \failcell{70.2} & \failcell{64.7} & \failcell{61.9} \\
\multirow{2}{*}{Manaca} & 1B & & \failcell{18.1} & \failcell{54.1} & \failcell{55.5} & \failcell{38.4} & \failcell{52.0} & \failcell{60.1} & \failcell{73.9} \\
 & 1B & \checkmark & \failcell{17.4} & \failcell{49.9} & \failcell{55.2} & \failcell{41.5} & \failcell{53.8} & \failcell{57.8} & \failcell{65.1} \\
\multirow{2}{*}{LLM-jp 3} & 13B & & \failcell{6.9} & \failcell{67.8} & \failcell{53.2} & \failcell{41.3} & \failcell{67.3} & \failcell{44.3} & \failcell{48.1} \\
 & 172B & & \failcell{6.0} & \failcell{64.5} & \failcell{49.5} & \failcell{34.8} & \failcell{65.9} & \failcell{41.0} & \failcell{41.8} \\
Swallow & 13B & & \failcell{6.2} & \failcell{66.2} & \failcell{50.7} & \failcell{28.9} & \failcell{72.6} & \failcell{37.8} & \failcell{54.4} \\
Chinese Llama 2 & 13B & & \failcell{4.4} & \failcell{62.9} & \failcell{54.1} & \failcell{24.3} & \failcell{70.0} & \failcell{63.6} & \failcell{41.2} \\
\bottomrule
\end{tabular}
\caption{Failure rates (\%) across languages, defined as the percentage of sentence pairs for which a model assigns a higher SLOR to the transferred strategy than to the native strategy. Darker red cells indicate higher failure rates. \checkmark{} in the IT column marks instruction-tuned models.}
\label{tab:failure_rates}
\end{table*}

\end{document}